\documentclass{article}
\usepackage{ijcai21}
\usepackage{multirow}
\usepackage{subcaption}
\usepackage{times}
\usepackage{soul}
\usepackage{url}
\usepackage[hidelinks]{hyperref}
\usepackage[utf8]{inputenc}
\usepackage{caption}
\usepackage{graphicx}
\usepackage{amsmath}
\usepackage{amsthm}
\usepackage{booktabs}
\usepackage{algorithm}
\usepackage{algorithmic}
\usepackage{enumerate}
\usepackage[american]{babel}
\usepackage{microtype} 
\title{Proposal-free One-stage Referring Expression via Grid-Word Cross-Attention}

\author{
    Paper ID 1319
    \affiliations
    Anonymous
}

\author{
Wei Suo$^{1}$\footnote{The first two authors equally contributed to this work. P. Wang is the
corresponding author. }\and
Mengyang Sun$^{1*}$\and
Peng Wang$^1$\And
Qi Wu$^2$
\affiliations
$^1$Northwestern Polytechnical University, China \\
$^2$University of Adelaide, Australia \\
\emails
\{suowei1994, sunmenmian\}@mail.nwpu.edu.cn, \\
peng.wang@nwpu.edu.cn, qi.wu01@adelaide.edu.au
}

\begin{document}

\maketitle

\begin{abstract}
Referring Expression Comprehension (REC) has become one of the most important tasks in visual reasoning, since it is an essential step for many vision-and-language tasks such as visual question answering. However, it has not been widely used in many downstream tasks because it suffers 1) two-stage methods exist heavy computation cost and inevitable error accumulation, and 2) one-stage methods have to depend on lots of hyper-parameters (such as anchors) to generate bounding box. In this paper, we present a proposal-free one-stage (PFOS) model that is able to regress the region-of-interest from the image, based on a textual query, in an end-to-end manner. Instead of using the dominant anchor proposal fashion, we directly take the dense-grid of image as input for a cross-attention transformer that learns grid-word correspondences. The final bounding box is predicted directly from the image without the time-consuming anchor selection process that previous methods suffer. Our model achieves the state-of-the-art performance on four referring expression datasets with higher efficiency, comparing to previous best one-stage and two-stage methods.
\end{abstract} 
\section{Introduction}
Referring Expression Comprehension (REC) aims to localize objects in images based on natural language queries. It is a fundamental building block in the field of human-machine communication, and it also can be applied to other vision-and-language tasks such as visual question answering \cite{gan2017vqs}, image retrieval \cite{chen2017amc} and visual dialogue \cite{zheng2019reasoning}. However, REC is difficult because this task requires a comprehensive understanding of complicated natural language and various types of visual information. There are mainly two kinds of approaches to address such challenges: two-stage approaches and one-stage approaches.
\begin{figure}[ht]
\includegraphics[width =0.49\textwidth]{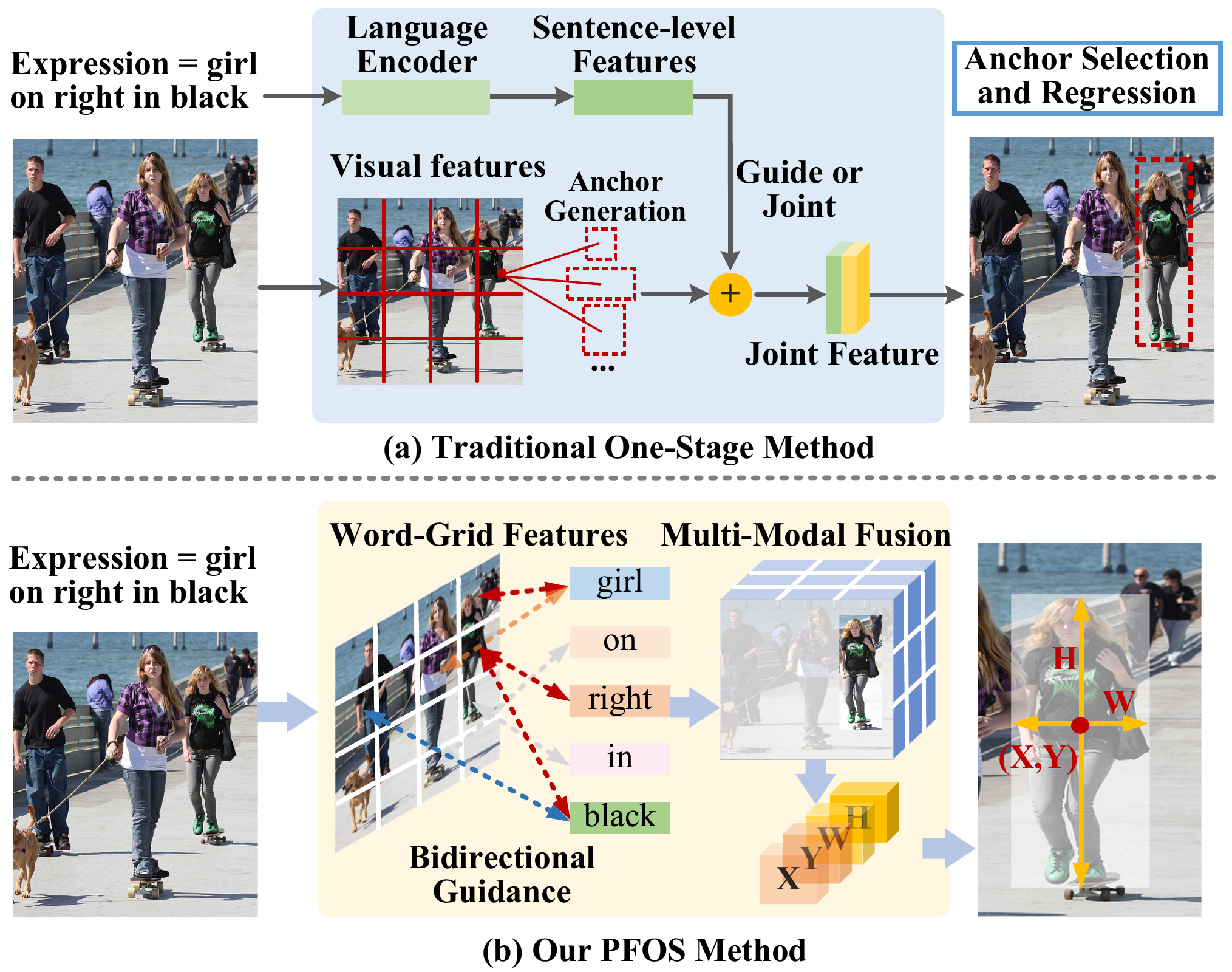}
\caption{A comparison of existing one-stage methods and our proposed method. (a) The traditional methods just leverage the sentence-level features to concatenate or guide the visual features, then select and regress anchors through complex computation. (b) Our model focuses on learning fine-grained grid-word correspondences and utilizes the multi-modal representations to directly produce the bounding box.}
\label{fig:first_img}
\end{figure}

Two-stage approaches usually treat REC as a ranking task \cite{zhang2018grounding,yu2018mattnet,wang2019neighbourhood}. They depend on mature object detectors such as Faster R-CNN to extract a set of candidate regions and a subtle model that is designed to select the best matching object. Although existing two-stage methods have achieved pretty-well performance, these frameworks are capped by the first stage with an inevitable error accumulation (if the target object can not be captured in the first stage, the frameworks fail no matter how good the second ranking stage could perform). On the other hand, \cite{yang2019fast} and \cite{liao2020real} has proved that two-stage methods exist heavy computation cost. For each object proposal, both feature extraction and cross-modality similarity computation have to be conducted.

As shown in Fig. \ref{fig:first_img} (a), different from two-stage approaches, one-stage approaches can directly predict bounding boxes and have the great advantage of speed and accuracy (avoid error accumulation). However, as a well-received computing paradigm, one-stage methods face two main challenges: 1) how to effectively model the relationship of both the visual and linguistic domains. 2) how to accurately generate the bounding box based on language. To address the above problems, existing one-stage methods usually utilize LSTM hidden state output for the last time step or special token ([CLS]) from BERT to encode the entire query as a single embedding vector. Then, the vector is concatenated with visual features \cite{yang2019fast} or treated as a kernel to perform correlation filtering \cite{liao2020real}. Although it is a convenient way to bridge the gap between vision and language, the sentence-level features might tend to miss some fine-grained information. Moreover, this type of unidirectional intervention causes models can not efficiently model correspondences between multi-modal. On the other hand, in order to accurately generate the bounding box, these approaches mostly need to pre-define anchors by K-means clustering over all the ground-truth boxes \cite{yang2020improving}. But the strategy of heuristic tuning would significantly hamper the generalization ability on the other datasets (more details in section \ref{sec:ablation}) and lead to complicated computation \cite{zhou2019objects}.

Intuitively, when we humans solve an REC problem, we tend to back-and-forth observe and extract the crucial information from the paired data, then we summarize these cues to arrive at final conclusion straightly. Inspired by this, we propose a novel proposal-free one-stage (PFOS) framework with cross-attention to solve the above problems in a better way. As shown in Fig.\ref{fig:first_img} (b), we first utilize the bidirectional guidance to model relationships between the visual and linguistic domains. In particular, our model depends on cross-attention module to learn fine-grained (grid-word) correspondences and passes inter-modal messages. Then, through the multi-modal fusion to further integrate these inferential cues and highlight the vital feature representations. Finally, different from previous one-stage methods that utilize anchor-based detection heads as their backbones, without bells and whistles, we introduce a simpler computing paradigm that only relies on the refined features to directly localize the region-of-interest. Our method completely avoids the selection of anchors and achieves faster inference speed than current methods.

To summarize, the main contributions of our paper are as follows:
\begin{itemize}
    \item [1)]
    Our proposal-free one-stage (PFOS) framework changes the unidirectional intervention into a bidirectional process with cross-attention. It can model multi-modal correspondences and learn more fine-grained feature representations than traditional one-stage methods.
    
    \item [2)]
    The PFOS can directly predict the bounding box from grid-word features that avoids redesigning anchors for different datasets and achieves faster inference than previous methods. 
    \item [3)]
    The PFOS achieves significant performance improvement over the state-of-the-art on four widely used referring expression datasets: ReferItGame, RefCOCO, RefCOCO+ and RefCOCOg.
\end{itemize}



\section{Related Work}
REC is the task of localizing the target object with a bounding box based on a given expression. Previous methods can be mainly divided into two types, including two-stage approaches and one-stage approaches.
\paragraph{Two-stage approaches.}
Most of the two-stage approaches \cite{yu2018mattnet,wang2019neighbourhood} reformulate REC as a ranking task with a set of candidate regions. Given an input image, these methods first generate proposal regions, which are typically extracted by a pre-trained object detector. In the second stage, the models align candidate regions with the expression, then select a target candidate that matches the corresponding expression. 
Though existing two-stage methods have achieved great success, their solutions have two notable issues. First, two-stage methods are very computationally
expensive. Second, these frameworks are capped by the performance of the first stage \cite{liao2020real}.

\paragraph{One-stage approaches.}Compared to two-stage approaches, one-stage methods can reach real-time processes and achieve superior grounding accuracy. 
\cite{yang2019fast} proposes a one-stage model that fuses a query’s embedding into YOLOv3, then it uses the merged features to localize the corresponding region. \cite{liao2020real} reformulates the REC as a correlation filtering process with CenterNet \cite{zhou2019objects}. The expression is treated as a kernel to perform correlation filtering on the image feature maps and the output is only used to predict the center. \cite{yang2020improving} leverages a recursive sub-query construction framework to reduce the referring ambiguity with complex queries.

However, these methods generally ignore the fine-grained pairwise interaction between words and image regions. Further, this type of unidirectional guidance from the language domain to image domain causes a semantic loss due to modality misalignment.
Different from previous methods, our method stacks cross-attention and self-attention modules to model pairwise word-grid interactions and fuses multi-modal information to generate the target bounding-box directly. 
\section{Propose Method}
\begin{figure*}[ht]
\centering
\includegraphics[width = 1\textwidth]{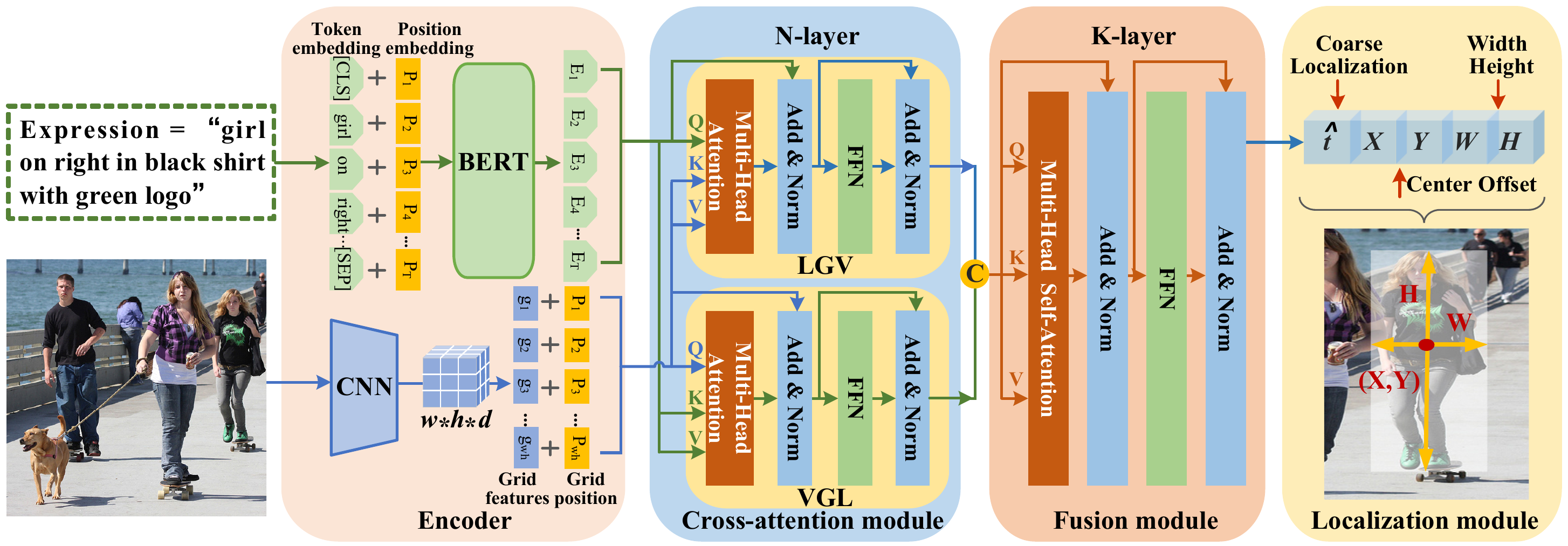}
\caption{The architecture for our proposal-free one-stage framework. It consists of three main parts: (a) Encoder module: BERT and CNN models are used for language and image features extraction, respectively. (b) Grid-word attention module: this module utilizes the bidirectional cross-attention module and fusion module to learn richly contextual vision-language representations. \includegraphics[width=0.32cm]{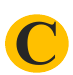}: sequence concatenation. (c) localization module: a simple anchor-free strategy that directly localizes the region-of-interest.}
\label{fig:model}
\end{figure*}
In this section, we introduce our proposal-free one-stage (PFOS) framework, a novel approach for referring expression comprehension. Our aim is to localize the object described by the query expression directly with image's grid-level features and expression's word-level features, without using any region proposals and prior anchors. Inspired by \cite{hu2020iterative,carion2020end} success at vision-language and object detection tasks, PFOS is an analogous to transformer-style model, as illustrated in Fig. \ref{fig:model}. But different from them, our model focus on jointly representing grid-level visual features and word-level text features from paired data. Moreover, it only depends on an encoder to directly produce bounding boxes through a simpler localization module. Next, we describe the components of our model in detail.
\subsection{Encoder Module}
Given an image $I\in R^{W\times H\times 3}$ and  a query expression sentence $S=\{s_t\}^T_{t=1}$, where $s_t$ represents the $t$-th word, and $W\times H\times 3$ denotes the size of the image. Our goal is to find one sub-region $I_S$ within the image $I$, which corresponds to the semantic meaning of the referring expression $S$.
\paragraph{Expression encoder.}In this paper, we use the uncased BERT as our expression encoder. First, each word in a referring expression is mapped to the corresponding word embedding vector. 
Then, each $s_t$ and its index $t$ ($s_t$’s absolute position in the sentence) is fed into query encoder (\textit{i.e.,} BERT). Following \cite{yang2020improving}, we sum the representation for each word in last four layers and each sentence also comes with special tokens such as [CLS], [SEP] and [PAD]. Finally, we obtain word-level features $E=\{e_t\}_{t=1}^T$, $e_t\in R^d$, where the dimension size $d=768$ and the maximum $T=20$.
\paragraph{Image encoder.}We follow \cite{yang2020improving} that uses Darknet-53 as the backbone to extract visual features for input images. Specifically, we first resize the given image $I$ to the size of $3\times 256\times 256$, then feed it into the encoder network. The output feature map is $G=\{g_i\}^{w\times h}_{i=1}, g_i\in R^{d_g}$, which denotes different local regions for the input image. In other words, the feature map spatial resolution is $w \times h$, and each $g_i$ represents a grid-level feature for the output feature map $G$. We add a $1\times 1$ convolution layer with batch normalization and RELU to map them all to the same dimension $d$.
\subsection{Grid-word Attention Module}
In this section, we first introduce the proposed cross-attention module and then describe how we apply these inferential cues to obtain refined feature representations from pairwise data.
\begin{table*}[ht]
\centering
\resizebox{\textwidth}{!}{
\begin{tabular}{ c  l  c  c  c  c  c c  c  c  c  c  c c}
\hline
\multirow{2}{*}{Type} & \multirow{2}{*}{Method} & \multirow{2}{*}{Backbone} & ReferItGame & \multicolumn{3}{c}{RefCOCO} & \multicolumn{3}{c}{RefCOCO+} & \multicolumn{3}{c}{RefCOCOg} \\
~ & ~ & ~ & test & val& testA & testB & val& testA & testB & val-g& val-u & test-u &time(ms)\\
\hline
\multirow{8}{15mm}{Two-stage Models}
~ & CMN\cite{hu2017modeling}	        &	VGG16	&	28.33	&	-	    &	71.03	&	65.77	&	-	    &	54.32	&	47.76	&	57.47	&	-	    &	-	   & -       \\
~ & VC\cite{zhang2018grounding}	        &	VGG16	&	31.13   &	-	    &	73.33	&	67.44	&	-	    &	58.40	&	53.18	&	62..30	&	-	    &	-	   & -       \\
~ & ParallelAttn\cite{zhuang2018parallel} 	&	VGG16	&	-	    &	-	    &	75.31	&	65.52	&	-	    &	61.34	&	50.86	&	58.03	&	-	    &	-	   &  -      \\
~ & LGRAN\cite{wang2019neighbourhood}        &	VGG16	&	-	    &	-	    &	76.60	&	66.40	&	-	    &	64.00	&	53.40	&	61.80	&	-	    &	-	   &  -      \\
~ & SLR\cite{yu2017joint}	        &	Res101  &	-	    &	69.48	&	73.71	&	64.96	&	55.71	&	60.74	&	48.80	&	-	    &	60.21	&	59.63  & -	    \\
~ & MAttNet\cite{yu2018mattnet}	    &	Res101	&	29.04	&	76.40	&	80.43	&	69.28	&	\underline{64.93}	&	\underline{70.26}	&	56.00	&	-	    &	66.67	&	67.01  & 320    \\
~ & DGA\cite{yang2019dynamic}	        &	Res101	&	-	    &	-	    &	78.42	&	65.53	&	-       &	69.07	&	51.99	&	-	    &	- 	    &	63.28  & 341       \\

\hline
\multirow{6}{15mm}{One-stage Models}
  & SSG\cite{chen2018real}	            &	Darknet53	&	54.24	&	-	    &	76.51	&	67.50	&	-	    &	62.14	&	49.27	&	47.47	&	58.80	&	-	    &   25   \\
~ & One-Stage-BERT\cite{yang2019fast}	&	Darknet53	&	59.30	&	72.05	&	74.81	&	67.59	&	55.72	&	60.37	&	48.54	&	48.14	&	59.03	&	58.70	&   23  \\
~ & RCCF\cite{liao2020real}	            &	DLA-34	    &	63.79	    &	-	    &	\underline{81.06}	&	71.85	&	-	    &	{\bf70.35}	&	56.32	&	-	    &	-	    &	65.73 &	25\\
~ & Sub-query-Base\cite{yang2020improving}	&	Darknet53	&	64.33	&	76.59	&	78.22	&	\underline{73.25}	&	63.23	&	66.64	&	55.53	&	60.96	&	64.87	&	64.87	&   26   \\
~ & Sub-query-Large\cite{yang2020improving}	&	Darknet53	&	\underline{64.60}	&	\underline{77.63}	&	80.45	&	72.30	&	63.59	&	68.36	&	\underline{56.81}	&	\underline{63.12}	&	\underline{67.30}	&	\underline{67.20}	&   36   \\
~ & {\bf PFOS(ours)}            	&	Darknet53	&	{\bf66.90}	&	{\bf79.50}	&	{\bf81.49}	&	{\bf77.13}	&	{\bf65.76}	&	69.61	&	{\bf60.30}	&	{\bf63.83}	&	{\bf69.06}	&	{\bf68.34} & {\bf 20} 	\\
\hline
\end{tabular}}
\caption{Comparison with the state-of-the-art methods on the ReferItGame, RefCOCO, RefCOCO+ and RefCOCOg datasets.}
\label{table:sota}
\end{table*}
\paragraph{Cross-attention module.}The cross-attention module is composed of two sub-modules: Language Guide Vision (LGV) and Vision Guide Language (VGL). 
Similar to \cite{tan2019lxmert}, each sub-module is composed of a stack of $N$ identical self-attention layers. We take LGV for example. The input consists of token embeddings $E=\{e_t\}_{t=1}^T$ and grid features $G=\{g_i\}^{w\times h}_{i=1}$ of dimension $d$.
We compute the dot products of the $e_t$ with all $g_i$, each of that is divided by $d'=\sqrt{d/m}$ ($m$ is the number of attention heads) and a softmax function is applied to obtain the attention weights on the $G$. For simplicity, we assume a single attention head in the LGV. Then, the attention in LGV can be formulated as:
\begin{equation}
    \begin{aligned}
    A^n_{LGV}& =f_{LGV}(E,G) =\mathrm{softmax}(\frac{QK^\mathsf{T}}{d'})V, \\
    Q&= W_Q^nE,K=W_K^nG,V=W_V^nG.
    \end{aligned}
\end{equation}
The variables $Q$, $K$ and $V$ indicate queries, keys and values as in the self-attention module and the embedding weights are denoted by $W_Q^n$, $W_K^n$, $W_V^n$ where $n$ represents the $n$-th layer of LGV module. Then the $A^n_{LGV}$ is further encoded by two feed-forward layers with a residual connection to form the output $H^n_{LGV}$. Since the transformer architecture is permutation-invariant, we follow \cite{carion2020end} that supplements grid features with spatial positional encoding. The VGL is similar to LGV, except that the G is encoded to queries and E is used as keys and values. Finally, we obtain two cross representations: $H_{LGV}\in R^{T\times d}$ and $H_{VGL}\in R^{{(w\times h)}\times d}$ by cross-attention module.
\paragraph{Fusion module.}In order to further establish the connection between grid-level visual features and word-level textual features, we adopt a multi-head fusion module which stacks $K$ layers of transformer-style model \cite{vaswani2017attention}. We concatenate the output of cross-attention module $H_{LGV}$ and $H_{VGL}$ to generate $H^0_F\in R^{(T+(w\times h))\times d}$, which is the input of the fusion module. 
After this fusion step, we only rely on the visual part of the last layer output $H_F\in R^{(w\times h)\times d}$ to directly predict the bounding box.
\subsection{Localization Module}
Different from the existing one-stage methods which utilize the anchor-based detection head, we introduce a simpler localization module in the field of REC that can localize the region of expression without processing anchors. 

Specifically, the output of above fusion module is passed through a convolutional layer which contains 5 filters by stride $1\times 1$ to obtain a feature map with a shape of $w \times h \times 5$ that represents five predicted values $\{\hat{t},t_x,t_y,t_w,t_h\}$. Assuming the target bounding box is $bbox=\{X_b,Y_b,W_b,H_b\}$, where $(X_b,Y_b)$ is the center point and $W_b,H_b$ are the width and height,  respectively.
Our center point prediction is carried out by two steps: coarse prediction and offset regression. For the width and height, we straightforwardly utilize the $(\frac{W_b}{W},\frac{H_b}{H})$ as groundtruth. The coarse prediction loss $L_{cls}$, offset of center loss $L_{off}$ and the regression loss $L_{rgr}$ are defined as:
\begin{align}
    &{L_{cls}}= -\sum_{i=1}^{w}\sum_{j=1}^hC_{ij}\mathrm{log}(\hat{t}_{ij})+(1-C_{ij})\mathrm{log}(1-\hat{t}_{ij}),\notag\\
    &{L_{off}}= (\Delta{x}-t_x)^2+(\Delta{y}-t_y)^2,\\
    &{L_{rgr}}= (\frac{W_b}{W}-t_w)^2+(\frac{H_b}{H}-t_h)^2,\notag
\end{align}
where $C_{ij}=$ 1 or 0 denotes whether the grid has the coarse center point. The offset of the center is $(\Delta{x},\Delta{y})=(\frac{X_b}{W/w}-\hat{x},\frac{Y_b}{H/h}-\hat{y})$, where $\hat{x}$, $\hat{y}$ denote $\mathrm{int}(\frac{X_b}{W/w},\frac{Y_b}{H/h})$ ( $\mathrm{int}(\cdot)$ operation round-down fractions to the closet integer). Note that the regression loss acts only at the location of coarse center point and we follow \cite{carion2020end} that uses $GIoU$ loss as our auxiliary loss. The final loss is the summarization of four loss terms:
\begin{equation}
{
    \begin{split}
        Loss={L_{cls}+\lambda_{off}{L_{off}}+\lambda_{rgr}{L_{rgr}}}+{L_{giou}},
    \end{split}
}
\end{equation}
Following \cite{yang2020improving}, where we set $\lambda_{off}$ and $\lambda_{rgr}$ to 5. At inference time, the network selects the point $(\hat{x},\hat{y})$ with the highest coarse score and straightforwardly generates the bounding box. The coordinates of the precise center point, width and height are obtained by:
\begin{equation}
{\small{
    \begin{split}
    (X_t,Y_t,W_t,H_t)=((\hat{x}+t_x)\frac{W}{w},(\hat{y}+t_y)\dfrac{H}{h}
    ,t_wW,t_hH).
    \end{split}
    }
}
\end{equation}
\section{Experiments}
\subsection{Experimental Setting}
\paragraph{Datasets.}We evaluate the proposed one-stage REC approach on four benchmarks including RefCOCO \cite{yu2016modeling}, RefCOCO+ \cite{yu2016modeling}, RefCOCOg \cite{mao2016generation} and ReferItGame \cite{kazemzadeh2014referitgame}. ReferItGame has 20,000 images from the SAIAPR-12 \cite{escalante2010segmented}. The other three datasets are all based on MSCOCO \cite{lin2014microsoft}. The RefCOCO consists of 142,210 referring expressions for 50,000 objects in 19,992 images. RefCOCO+ is similar to RefCOCO but forbids using absolute location words, so it takes more attention on appearance. RefCOCOg contains 104,560 referring expressions for 54,822 objects in 26,711 images. It was collected in a non-interactive setting thereby producing longer expressions than that of the other three datasets which were collected in an interactive game interface.
\paragraph{Implementation details.}Following \cite{yang2019fast}, we keep the original image ratio when we resize an input image. We resize its long edge to 256 and then pad the resized image to $256\times 256$ with the mean pixel value. We follow \cite{yang2020improving} for data augmentation. The proposed PFOS's cross-attention module and fusion module are set to two and four layers respectively. We utilize Adam \cite{kingma2014adam} as our optimizer and the batch size is set to 8. The number of heads and the size of the feature map are 8 and $16\times 16 \times 512$.  We train on one 1080Ti GPU for 100 epochs with an initial learning rate of 5e-5, which is dropped by half every 10 epochs.
\paragraph{Evaluation.}The predicted region is considered as the right grounding if the Intersection-over-Union(IoU) score is greater than 0.5 with the ground-truth bounding box.
\subsection{Quantitative Evaluation}
As illustrated in Table \ref{table:sota}, in the top part, we quote two-stage methods' results reported by \cite{yang2020improving} and compare these on ReferItGame, RefCOCO, RefCOCO+ and RefCOCOg. 
Note that although these two-stage methods also achieve good performance, they have to depend on object detectors to extract candidate regions with additional supervision (attributes or class labels).Moreover, typical two-stage approaches also take lots of time to process pairwise proposal-query information, such as MattNet \cite{yu2018mattnet} needs to cost 320ms for a sample. However, our PFOS is about 16-times faster than the state-of-the-art two-stage method with higher accuracy.
\begin{table}[ht]
\centering
\resizebox{0.49\textwidth}{!}{
\begin{tabular}{l c c c }
\toprule[1pt]
\multirow{2}{*}{Method}	& ReferItGame & \multicolumn{2}{c}{ReferCOCO}  \\	
~ & test & testA & testB 	 \\
\hline
Concatenation\cite{yang2019fast}         &     59.30     & 74.81      &    67.59    \\
PFOS:LGV                                      &     62.05     & 77.13      &	71.36	 \\
PFOS:VGL	                                  &     59.67     & 78.47      &	72.15	 \\
PFOS:LGV+VGL	                              &     64.60     & 80.75      &	75.49	 \\
PFOS:Fusion                                   &     65.79     & 81.12      &	75.28	 \\
\hline
PFOS:LGV+VGL+Fusion                           &     {\bf66.90}    & {\bf81.49}      &	{\bf77.13}	 \\
\bottomrule[1pt]
\end{tabular}}
\caption{Ablation experiments on the ReferItGame and RefCOCO datasets.}
\label{table:ablation}
\end{table}
\begin{table}
\centering
\resizebox{0.48\textwidth}{!}{
\begin{tabular}{c c c c c c}
\toprule[1pt]
\/PFOS & Attention &\multirow{2}{*}{Anchor} &ReferItGame &\multicolumn{2}{c}{ReferCOCO} \\	
Backbone & heads & ~  & test & testA & testB	 \\
\hline
ResNet50	&	8	&	-	 &65.82    	 &	80.61	& 72.05\\
ResNet101	&	8	&	-	 &66.28	     &	80.82   & 72.87	\\
\hline
Darknet53	&	4	&	-    &66.10 	 &	81.33   & 76.21	\\
Darknet53	&	12	&	-	 &66.73   	 &	81.05   & 75.88	\\
\hline
Darknet53	&	8	&	ReferitGame &66.33  &	77.95   & 74.63	\\
Darknet53	&	8	&	ReferCOCO   &62.99 &	79.88   & 75.93	\\
\hline
Darknet53	&	8	&	-          &{\bf66.90}	&	{\bf81.49} &{\bf77.13}	\\
\bottomrule[1pt]
\end{tabular}}
\caption{The effects of different experimental details on the ReferItGame and ReferCOCO dataset.}
\label{table:compare}
\end{table}

In the bottom portion of Table \ref{table:sota}, we compare the performance of our method with the best one-stage methods. We observe that our PFOS method uses fewer hyper-parameters and a much simpler structure to achieve better performance than the previous methods \cite{yang2020improving,liao2020real}. Overall, the PFOS can outperform the approaches whatever one-stage or two-stage on four benchmarks except for the result on RefCOCO+ testA that is slightly inferior to RCCF \cite{liao2020real}.
\subsection{Ablation Studies}
\label{sec:ablation}In this section, we conduct several ablation studies on the ReferItGame and RefCOCO datasets to demonstrate the effectiveness of our method.
\paragraph{Influence of main modules.}First, we study the benefits of each module of PFOS. The One-Stage-BERT \cite{yang2019fast} is used as our baseline model which simply concatenates visual and linguistic features to generate multi-modal representations. In Table \ref{table:ablation}, we follow the baseline to concatenate the output of the LGV or VGL and features extracted by the encoder. The results show that LGV or VGL is available as an effective sub-module and combining these two sub-modules boost the baseline accuracy by 7.90\% on the RefCOCO testB. Meanwhile, employing fusion module outperforms the baseline by 7.69\%. Furthermore, when we integrate the cross-attention module and fusion module together, the performance improves to 77.13\% on the RefCOCO testB over the baseline model by 9.54\%.
\paragraph{Different backbones, heads and anchors.}
\begin{figure}[t]
\begin{subfigure}{0.235\textwidth}
\includegraphics[width = 1\textwidth,height=4cm]{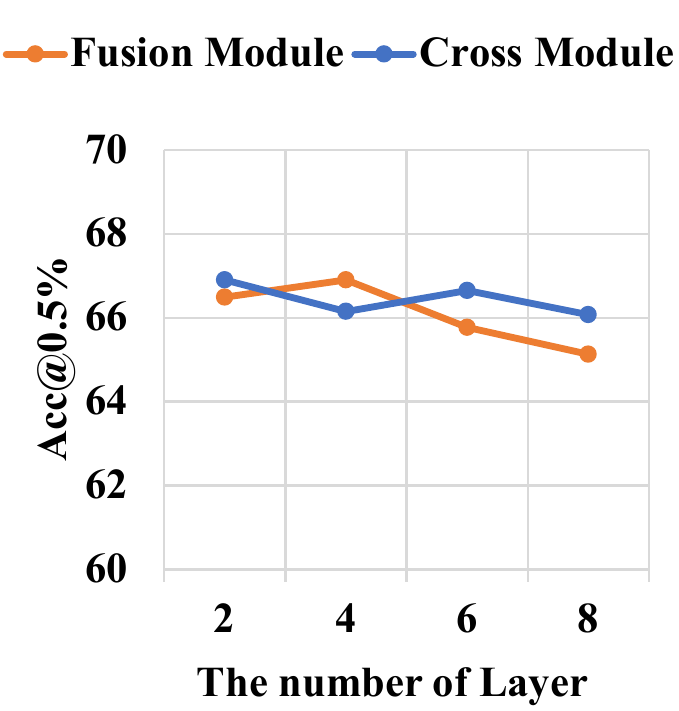}
\caption{} 
\end{subfigure}
\begin{subfigure}{0.235\textwidth}
\includegraphics[width = 1\textwidth,height=4cm]{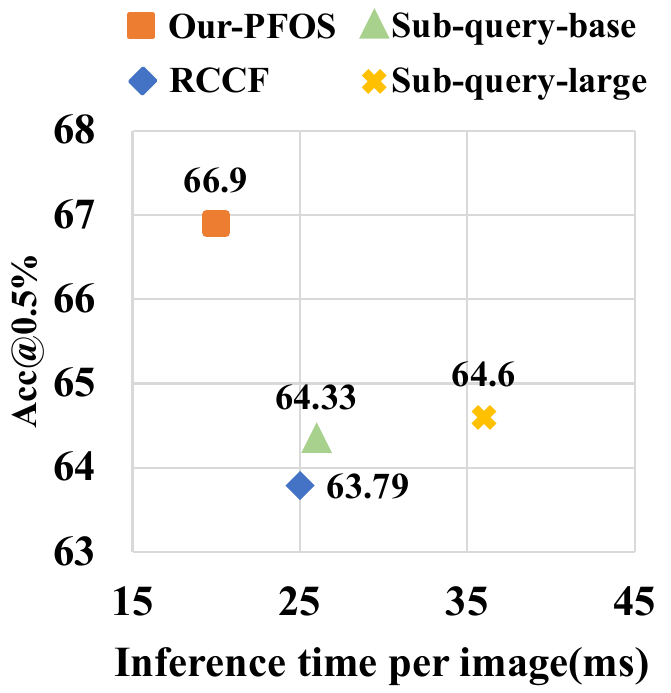}
\caption{}
\end{subfigure}
\caption{(a) The effects of different number of layers in fusion module and cross-attention module on the ReferItGame. (b) The inference time ($ms$ per sample) and performance comparisons on the ReferItGame dataset between our PFOS and other one-stage methods.}
\label{fig:speed}
\end{figure}

In Table \ref{table:compare}, we first compare the performance of our method with different CNN backbones. Because Darknet-53 is pre-trained on the COCO for object detection, it may provide prior knowledge about the datasets. Here we show results on the Resnet-50 and Resnet-101 \cite{he2016deep} pre-trained on ImageNet for image classification in the first two rows. We find that we can also obtain a competitive performance compared to SOTA models on the ReferCOCO even without any in-domain information. Moreover, there are no obvious changes in the performance of our PFOS on the ReferItGame. 
The above results prove that the performance improvement of PFOS does not only rely on the prior knowledge or the structure of the visual encoder. In row 3 and 4 of Table \ref{table:compare}, we find that our model is similar to \cite{vaswani2017attention} and the quality also drops a little with too many heads. In the last three rows of this table, we explore whether differently prior anchors would heavily influence the results. We follow \cite{yang2020improving} that utilizes K-means clustering over all the ground-truth boxes to generate the pre-define anchors. Then YOLOv3's detection head is used to replace our localization module. As we predicted, if we use the anchors from ReferCOCO instead of ReferItGame, then the results would drop about 3\% on the ReferItGame. We can also find the same trend on the ReferCOCO. Instead, our localization module can completely avoid the severe decline with anchor-free. 
\paragraph{Different layers.} 
Intuitively, the cross-attention module attends to the crucial information in visual and language. While the fusion module performs a more complicated reasoning process. In order to investigate the impact of the different number of layers on the performance of PFOS, as shown in Fig. \ref{fig:speed} (a). The Blue line indicates that fixing the fusion module layers to four and change the number of cross-attention module layers. Instead, we fix the cross-attention module’s layers to two and change the other one as shown in the orange line. We find that more layers are unable to achieve an obvious improvement in performance.
\paragraph{Speed comparisons.}
\begin{figure*}[ht]
\centering
\includegraphics[width = 1\textwidth]{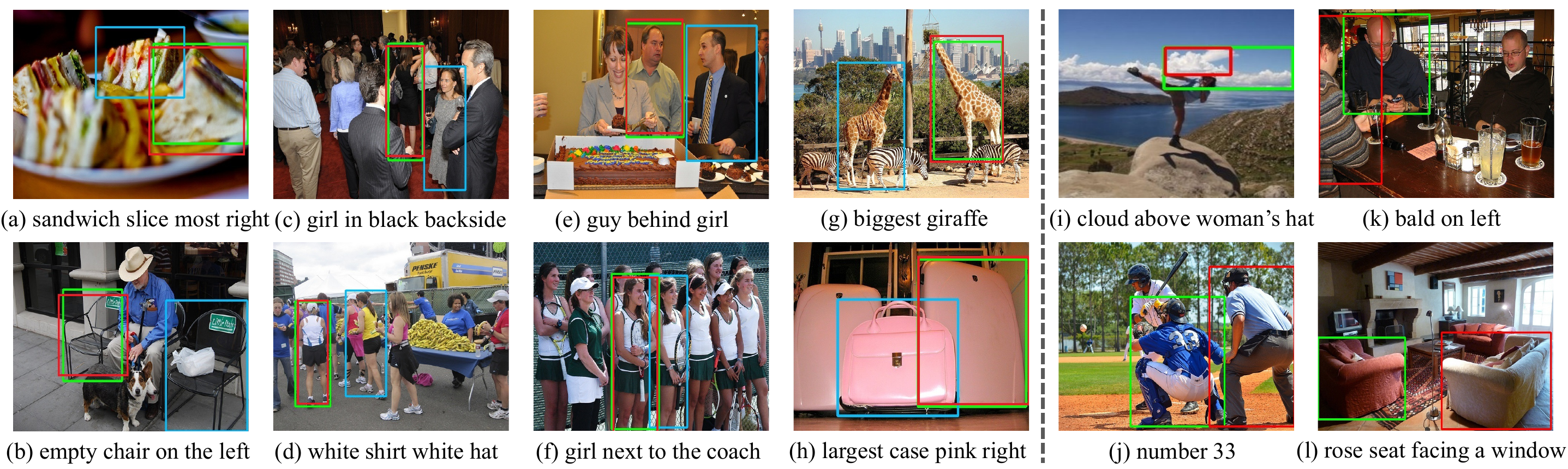}
\caption{The left four columns show success cases of our one-stage method. Green/blue boxes are ground-truths/predicted regions by the state-of-the-art one-stage method, while the red ones denote predicted boxes by our PFOS. Our approach performs better on four types of queries referring: (a,b) referring to an absolute location, (c,d) referring to attributes, (e,f) relative position of objects, and (g,h) comparing between objects. The last two columns are some failures in our method.}
\label{fig:visualization}
\end{figure*}
\begin{figure*}[ht]
\includegraphics[width = 1\textwidth]{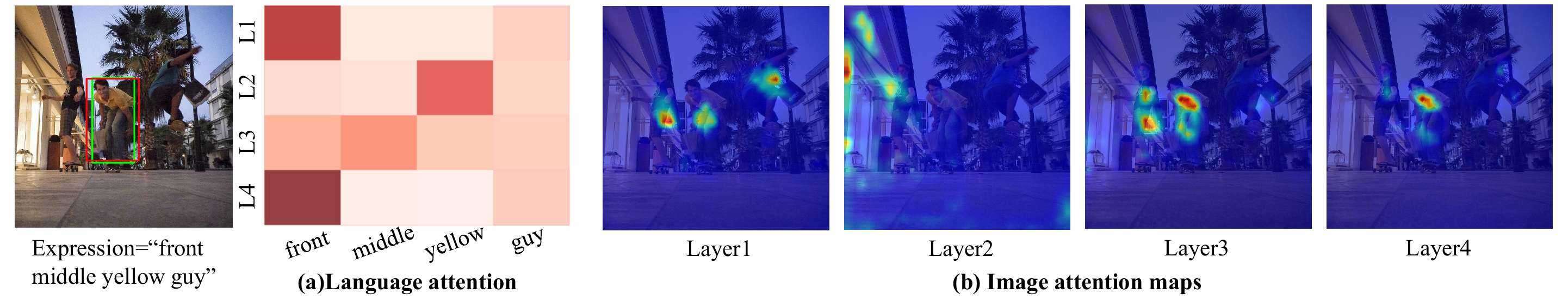}
\caption{Visualizations of the heatmaps of the fusion module at each layer. More examples are in supplementary materials.}
\label{fig:attention}
\end{figure*}
Finally, as shown in Fig. \ref{fig:speed} (b), we compare the inference time with mainstream one-stage methods on a desktop with a single 1080Ti GPU and Intel Xeon CPU Gold 5118@2.3G. The results show that PFOS can outperform other methods in both accuracy and speed with a simpler structure and fewer hyper-parameters. It also means that the proposed model possesses a better ability to learn multi-modal representations and can be applied to practical scenarios with real-time requirements. Specifically, the inferential time decreases by about 20 percent compared to SOTA methods and the accuracy is increased by around 2.5\%.

\subsection{Qualitative Results Analyses}

In order to analyze advantages and weaknesses of the proposed method, Fig. \ref{fig:visualization} presents some qualitative visual results on RefCOCO testA and testB. The left four columns show the success cases of our proposed model but failure by the previous one-stage method. The green boxes are ground-truths, the blue boxes are predicted by the state-of-the-art method \cite{yang2020improving} and the red ones are predicted by our method.
 
Fig. \ref{fig:visualization} (a) and (b) query the absolute position of objects in the image. (c) and (d) show examples of referring regions that are selected from multiple objects according to some attributes such as color and accessories in images. (e) and (f) provide query expressions that contain at least two objects. Therefore, the model is required to judge the relative position relationship between several objects. Additionally, Fig. \ref{fig:visualization} (g) and (h) show our method is more competitive in reasoning about comparative relationships.\par

From the above success cases, we find that our model is more sensitive to attributes and relationships between multiple objects than the previous methods. 
In order to better explore the reasoning processes learned by the PFOS, we study the visualizations of sample results along with the attention distuibution produced by the PFOS. At each layer of the fusion module, we show the language attention over the words and attention distribution over the center point which is predicted by our coarse localization.
In Fig. \ref{fig:attention} (a), L1 to L4 represent the layers of our model. It can be observed that ‘front’ and ’guy’ have a relatively large attention weight in L1. Accordingly, the key regions are positioned in Fig. \ref{fig:attention} (b) first attention map for the “front guy”. Meanwhile, the model can focus on important attributes such as “middle” and “yellow” to distinguish similar objects. 
It follows that our model is able to attend to the right regions even though it is learned without any additional information such as objects' position and attributes. 

Finally, the right two columns of Fig. \ref{fig:visualization} show some failure cases of our method. In Fig. \ref{fig:visualization} (i) the referring expression cause ambiguity which leads to the IoU between prediction and the ground-truth region is less than 50\%. Besides, most errors are due to lacking OCR ability, rare attributes in training data, and complex challenging expression understanding, for instance, in Fig. \ref{fig:visualization} (j), (k) and (l), respectively.
\section{Conclusion}We present the PFOS, an innovative and efficient one-stage method for referring expression comprehension. Different from the previous one-stage or two-stage models, 
the PFOS directly predicts the bounding box from the image and language query with higher accuracy. Furthermore, the dense-grid and word features enter into the cross-attention transformer straightly to learn fine-grained grid-word correspondences. The experimental results demonstrate that PFOS achieves better performance on both accuracy and speed.
\newpage
\bibliographystyle{named}
\bibliography{ref}
\end{document}